%% file: eacl2021.tex
\documentclass[11pt,a4paper]{article}
\usepackage[hyperref]{eacl2021}
\usepackage{times}
\usepackage{latexsym}

\usepackage{makecell,booktabs}
\usepackage{enumitem}
\usepackage{float}
\usepackage{xspace}
\usepackage{color}
\usepackage{pgf,tikz,pgfplots}
\pgfplotsset{compat=1.16}

\newcommand{\tablecolor}[1]{\textcolor{red}{#1}}
\newcommand{\errorcolor}[1]{\textcolor{blue}{#1}}
\newcommand{\Sref}[1]{\S\ref{#1}}

\newcommand{\Fref}[1]{Figure~\ref{#1}}

\newcommand{\Tref}[1]{Table~\ref{#1}}
\newcommand\aspace{\hspace{.75em}}
\usepackage{colortbl}
\restylefloat{table}
\usepackage{microtype}
\usepackage{subcaption}
\usepackage{amsmath}
\usepackage{mathtools}
\usepackage{subcaption}
\usepackage{sidecap}
\sidecaptionvpos{figure}{t}

\newcommand\rurl[1]{%
  \href{https://#1}{\nolinkurl{#1}}%
}

\usepackage{microtype}

\aclfinalcopy %
\def\aclpaperid{1122} %

\newcommand\mypar[1]{\smallskip\noindent\textbf{#1}\quad}
\title{NoiseQA: Challenge Set Evaluation for User-Centric Question Answering}
\author{Abhilasha Ravichander \aspace\aspace Siddharth Dalmia \aspace\aspace Maria Ryskina \\
\textbf{Florian Metze} \aspace\aspace \textbf{Eduard Hovy} \aspace\aspace \textbf{Alan W Black} \\
Language Technologies Institute, Carnegie Mellon University, USA \\
\texttt{\{aravicha,sdalmia,mryskina\}@cs.cmu.edu} \\
}

\date{}

\begin{document}
\maketitle
\begin{abstract}
When Question-Answering (QA) systems are deployed in the real world, users query them through a variety of interfaces, such as speaking to voice assistants, typing questions into a search engine, or even translating questions to languages supported by the QA system. While there has been significant community attention devoted to identifying correct answers in passages assuming a perfectly formed question, we show that components in the pipeline that precede an answering engine can introduce varied and considerable sources of error, and performance can degrade substantially based on these upstream noise sources even for powerful pre-trained QA models. We conclude that there is substantial room for progress before QA systems can be effectively deployed, highlight the need for QA evaluation to expand to consider real-world use, and hope that our findings will spur greater community interest in the issues that arise when our systems actually need to be of utility to humans.\footnote{All resources available at
\rurl{noiseqa.github.io}}

\end{abstract}

\section{Introduction}
 Everyday users now benefit from powerful QA technologies in a range of consumer-facing applications including  
 health \cite{jacquemart2003towards, luo2015simq, Abacha2016RecognizingQE, kilicoglu2018semantic, guo2018qcorp}, 
 privacy \cite{sathyendra2017helping, harkous2018polisis, ravichander-etal-2019-question}, 
 personal finance \cite{alloatti-etal-2019-real}, 
 search \cite{yang-2015-deep, bajaj2016ms, he-etal-2018-dureader, kwiatkowski-etal-2019-natural} 
 and dialog agents \cite{dahl-etal-1994-expanding, Raux2005LetsGP}.  Voice assistants such as Amazon Alexa\footnote{\rurl{developer.amazon.com/alexa}} or Google Home\footnote{\rurl{assistant.google.com}} have brought natural language technologies to several million homes globally~\cite{osborne2016100, jeffs2018ok}. Yet, even with millions of users now interacting with these technologies on a daily basis, there has been surprisingly little research attention devoted to studying the issues that arise when people use QA systems.
 
\begin{table*}[tb]
\small
\setlength\tabcolsep{2pt}
\resizebox{1.0\linewidth}{!}{
\begin{tabular}{p{0.28\textwidth} p{0.1\textwidth} 
p{0.3\textwidth} p{0.3\textwidth}}
 \\[-1.75ex] \\ Original Question & \multicolumn{1}{c}{Interface} & Synthetic Construction & Natural Construction  \\ \midrule
What has a Lama determined to do? & \multicolumn{1}{c}{\begin{minipage}{.13\columnwidth}
      \includegraphics[width=\textwidth]{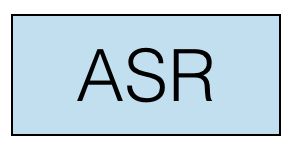}
  \end{minipage}} & what has a llama determined to do & what has a llama determined to do \\ \midrule
What has a Lama determined to do? & \multicolumn{1}{c}{\begin{minipage}{.13\columnwidth}
      \includegraphics[width=\textwidth]{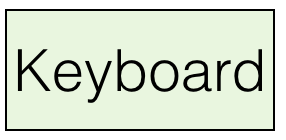}
  \end{minipage}} & Wjat has a Lsma determined yo do?  & WHat has a Lama determied to do?\\  \midrule
What has a Lama determined to do? & \multicolumn{1}{c}{\begin{minipage}{.13\columnwidth}
      \includegraphics[width=\textwidth]{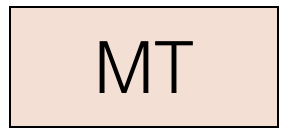}
  \end{minipage}} & What has a Lama decided to do? & What is a llama determined to do? \\ \bottomrule
\end{tabular}
}
\caption{Example question perturbations from synthetic and natural noise challenge sets for three types of interfaces: Automatic Speech Recognition (ASR) systems, Keyboard and Machine Translation (MT) systems.}
\label{tab:dataex-all}
\end{table*}
Traditional QA evaluations do not reflect the needs of many users who \emph{can} benefit from QA technologies. For example, users with a range of visual  and motor impairments now rely extensively on voice interfaces \cite{Pradhan2018AccessibilityCB} 
for efficient text entry.\footnote{More than 3.4 million American adults over the age of 40 have a form of visual impairment \cite{congdon2004causes}.} Another need is cross-lingual information access, e.g.\ in scenarios where a speaker of one of the $\sim$7000 non-English living languages in the world~\cite{lewis2009ethnologue} may want to take advantage of an English QA system.\footnote{As of 2021-01-24, there are 6,235,415 articles on English Wikipedia making it the largest edition: \rurl{wikicount.net}}
QA evaluation has to keep up with the different ways in which users may use these systems in practice, and the different users who interact with these systems. 

Keeping these needs in mind, we construct evaluations considering the \emph{interfaces} through which users interact with QA systems.\footnote{`QA system refers to any computing engine that receives a users' question and constructs an answer. It may consist of an end-to-end neural architecture or a structured pipeline.} We analyze errors introduced by three interface types that could be connected to a QA engine: 
{\bf speech recognizers} converting spoken queries to text, 
{\bf keyboards} used to type queries into the system, and 
{\bf translation systems} processing queries in other languages. Our contributions are as follows:

\begin{enumerate}[itemsep=1pt,topsep=2pt,parsep=2pt]
    \item We identify and describe the problem of interface noise for QA systems. We construct a challenge set framework for errors introduced by three kinds of interfaces: speech recognizers, keyboard interfaces, and translation engines, based on the popular SQuAD question-answering benchmark \cite{rajpurkar-etal-2016-squad}.  We define synthetic noise generators, as well as manually construct natural noise challenge sets, by processing SQuAD questions through the specified interfaces.
    \item  We evaluate the performance of current state-of-the-art methods on natural and synthetic noisy data. We find that accessibility needs to be consciously worked towards, as we see that the performance of QA systems can be impacted by the choice of interface.
    \item We analyze the generated noise and its impact on the downstream question answering and conduct an initial exploration of mitigation strategies for interface errors, focusing on data augmentation and query repair.
\end{enumerate}

\section{Motivation}
\label{sec:noise}

Modern QA systems often rely on large databases of digital text such as Wikipedia as their source of knowledge; such corpora typically contain well-formed text in a high-resource language like English. However, the user's input could come in many different forms: it could be spoken, or written but in another, possibly lower-resource language. To convert these inputs into the format that the system can process, another machine learning system such as a speech recognizer or a machine translation engine is required, and these intermediate systems will inevitably propagate their decoding errors into the QA engine. However, interface errors are not necessarily artifacts of machine learning models: even when the question comes in the desired form (e.g.\ English text), it has to be communicated to the QA system through a mechanical interface such as a keyboard, and the process of typing can introduce errors such as character substitutions. To be useful in real-world settings, a QA system has to be able to correctly process the input question regardless of the input interface. We simulate the use cases for three interface categories (ASR, MT, and keyboard) with different level of human involvement, from fully automatic pipelines to leveraging existing human-generated resources to  manual annotation, and evaluate whether the modern QA systems are capable of going from controlled well-formed inputs to real-world scenarios.

\section{Challenge Set Construction}

We define a suite of three types of noise perturbations, each imitating noise specific to a category of interfaces, and apply them to the data to create the challenge sets. 
We choose to add the noise to the questions but not to the context paragraphs, to replicate a realistic scenario of the noise being introduced to the question by the interface through which the user interacts with the QA engine. 
For each type of noise, we both build a synthetic generator that can introduce noise on a large scale, as well as manually create `natural' noise challenge sets to imitate real-world noise.

Our challenge sets are based on SQuAD~1.1~\cite{rajpurkar-etal-2016-squad},\footnote{Though in principle, these constructions could be applied to any kind of QA dataset} %
a large-scale machine comprehension dataset based on Wikipedia articles where the answer to each question is a span in a provided context. We choose SQuAD both for its popularity as a benchmark~\cite{gardner-etal-2018-allennlp, devlin2018bert, radford2018improving, wolf2019huggingfaces} and to avoid additional confounds such as unanswerable questions~\cite{rajpurkar-etal-2018-know}.\footnote{Future work would pursue a context-driven evaluation of unanswerability, identifying the kinds of unanswerable questions users ask in practice~\cite{ravichander-etal-2019-question,asai2020challenges}.} We use the standard $\sim$90K/10K train/development split and construct the challenge sets from the XQuAD data~\cite{artetxe-etal-2020-cross}, a subset of 1,190 SQuAD development set questions accompanied by professional translations into ten languages.\footnote{Spanish, German, Greek, Russian, Turkish, Arabic, Vietnamese, Thai, Chinese, and Hindi.} Below we discuss each challenge set in more detail.

\subsection{MT Noise}
Our first challenge set emulates machine translation noise introduced when the question is asked in a language other than the language of the QA system's training data. %
We use English as the QA system language, pairing English contexts with non-English questions.

\mypar{Synthetic Challenge Set}
Our synthetic noise generator employs the back-translation technique \cite{sennrich-etal-2016-improving, dong-etal-2017-learning, yu2018qanet}. 
In our case, back-translation is not meant to act as a data augmentation technique but rather to simulate noise that could be introduced by an MT engine when translating the question from another language. 
We imperfectly approximate natural non-English input by automatically translating English questions into a pivot language (German); we then translate them back to English, imitating a scenario where the user submits a query through an MT engine. 
We use the HuggingFace implementation~\cite{wolf2019huggingfaces} of MarianNMT \cite{mariannmt}.\footnote{
\href{https://huggingface.co/Helsinki-NLP}{\nolinkurl{huggingface.co/Helsinki-NLP/opus-mt-}}\texttt{\{}\href{https://huggingface.co/Helsinki-NLP/opus-mt-en-de}{\nolinkurl{en-de}}\texttt{|}\href{https://huggingface.co/Helsinki-NLP/opus-mt-de-en}{\nolinkurl{de-en}}\texttt{\}}
}

\mypar{Natural Challenge Set}
To bring our simulation closer to the natural setting, we create another challenge set from English machine translations of human-generated questions in other languages. We take the questions from the XQuAD dataset, which consists of English questions paired with professional translations into ten other languages.\footnote{A subtle nuance is that XQuAD questions are not originally written in these languages but translated from English; acknowledging this, we use XQuAD data as the natural challenge set because its fully parallel nature allows varying input language while controlling for content for fair comparison.} 
For each of the test set languages, we use Google's commercial translation engine\footnote{\rurl{translate.google.com}} to produce the English translation of the question. This allows us to construct ten challenge sets of translations from different languages with 1,190 questions each.

\subsection{Keyboard Noise}
\label{sec:keyboard}
This challenge set represents the noise introduced in the process of typing a question up on a keyboard, for example, when a question is submitted to a QA system through a search engine.

\mypar{Synthetic Challenge Set} Inspired by prior work~\cite{belinkov2017synthetic, naik-etal-2018-stress}, our basic noise generator introduces per-character typos based on the proximity of the keys in a standard QWERTY keyboard layout. Each word is corrupted with a 25\% probability by substituting a randomly sampled character with its row-wise neighbor. 
We also create more natural-looking noise by introducing externally collected human misspellings into our data on word level, as proposed by~\citet{belinkov2017synthetic}. Although prior work refers to this as natural noise, emphasizing that the typos have been produced by humans, we consider it synthetic because the errors are applied to the data outside of their original context. We start with the Wikipedia common English misspellings list\footnote{\rurl{en.wikipedia.org/wiki/Wikipedia:Lists_of_common_misspellings}} 
and apply a simple filtering heuristic that only retains keyboard errors (see Appendix~\ref{sec:appendix_epitran}), obtaining 1,742 misspellings for 1,489 English words.

\mypar{Natural Challenge Set}
To 
generate errors specific to the context of the question rather than hypothesized to exist at a lexical level across contexts, we ask three human annotators to retype English XQuAD questions. Annotators can see the original question, which helps avoid errors caused by misconception (e.g.\ not knowing the correct spelling of a named entity), but not their own input, in order to prevent them from correcting the typos. Of the obtained noisy questions, 51.6\% and 25.7\% differ from the original by at least one or at least two characters respectively.

\subsection{ASR Noise}
\label{sec:asr_noise_section}
Our final challenge set simulates ASR errors that occur when a question is posed to a voice interface.

\begin{table}[tb]
\small
\setlength\tabcolsep{2pt}
\begin{tabular}{p{1.5cm} p{6cm}}
\toprule
\textsc{Original Question} & How many Panthers defense players were selected for the Pro Bowl?\\
\midrule
\textsc{Google ASR} & how many Santa's defense players selected for the Pro Bowl\\ 
\midrule
\textsc{ESPnet (with LM)} & how many pantols the tent places were slected for the probol \\
\midrule
\textsc{Kaldi (with LM)} & how many friends tons of defence UNK for the UNK \\
\bottomrule
\end{tabular}
\caption{Example outputs of different ASR systems on a recorded question from SQuAD~\cite{rajpurkar-etal-2016-squad}.}
\label{tab:dataex-asr}
\end{table}

\mypar{Synthetic Challenge Set} 
We emulate automatic recognition of natural speech by using a Text-to-Speech (TTS) system pipelined with an ASR engine~\cite{tjandra2017listening}. 
We voice the questions using Google TTS and transcribe the obtained speech using Google Speech-to-Text optimized for English--US.
Besides Google ASR, we use Kaldi ASpIRE~\cite{povey2011kaldi, peddinti2015jhu} and ESPnet CommonVoice~\cite{watanabe2018espnet, ardila-etal-2020-common} open-source systems, as shown in Table \ref{tab:dataex-asr}.
We choose the former for analyzing the downstream effect of out-of-vocabulary word prediction in fixed vocabulary decoding~\cite{peskov2019mitigating}
and the latter for data augmentation (\Sref{sec:mitigation}) due to its improved out-of-vocabulary word handling with subword units.
In order to generate the large amount of speech data needed for augmentation, we use the open-source ESPnet LJSpeech TTS~\cite{hayashi2020espnet, ljspeech17} to voice the questions.

\mypar{Natural Challenge Set}
We use the SANTLR speech annotation toolkit \cite{li2019santlr} to record spoken versions of the prompt question from three human annotators (for background details, see Appendix~\ref{sec:appendix_asr}).
The obtained recordings are then transcribed using the ASR engines listed above. As expected, recognizing human speech is more difficult: %
the word error rate of the Google ASR system on the obtained set is 31\%, compared to 17\% on the synthesized English--US speech.

\definecolor{light-gray}{gray}{0.94}
\begin{table}[t]
\centering
\small
\resizebox{\columnwidth}{!}{
\setlength\tabcolsep{2pt}
\begin{tabular}{l @{\hskip 1.5em} c @{\hskip 1.5em} c @{\hskip 1.5em} c} \\ \toprule
Interface & CER ($\downarrow$) & WER ($\downarrow$) & BLEU ($\uparrow$)\\
\multicolumn{4}{c}{\cellcolor{light-gray} Synthetic}\\
ASR & \hphantom{0}3.96 & 16.61 & 77.12\\ 
Keyboard & \hphantom{0}4.11 & 23.93 & 52.66\\ 
Translation & 20.51 & 29.36 & 58.42\\ 
\multicolumn{4}{c}{\cellcolor{light-gray} Natural}\\
ASR & 12.96 & 30.67 & 57.22\\ 
Keyboard & \hphantom{0}1.78 & \hphantom{0}7.42 & 85.78\\ 
Translation & 31.89 & 43.34 & 47.07\\ 
 \bottomrule
\end{tabular}
\caption{\% Character Error Rate (CER), \% Word Error Rate (WER) and BLEU scores for all challenge sets compared to ground truth. 
For ASR and MT, synthetic noise is less prominent than natural, reflecting the idealized simulation conditions.
As expected, natural keyboard noise demonstrates the best word-level statistics.}
\label{tab:cerwer}
}
\end{table}

\section{Experiments}
We select four QA models that demonstrated strong performance on SQuAD~1.1\footnote{F1 scores on SQuAD dev set: BiDAF: 77.8; BiDAF-ELMo: 80.7; BERT: 88.8; RoBERTa: 89.9. For hyperparameters and implementation details, see Appendix~\ref{sec:appendix_hyperpar}.} to be tested under interface distortions: BiDAF~\cite{seo2016bidirectional}, which represents contexts at different levels of granularity using bidirectional attention flow mechanism; its extension BiDAF-ELMo~\cite{peters-etal-2018-deep} augmented with contextualized embeddings; BERT~\cite{devlin2018bert}, a bidirectional Transformer-based language model \cite{vaswani2017attention}; and  RoBERTa~\cite{liu2019roberta}, a more robustly pre-trained version of BERT.

\definecolor{light-gray}{gray}{0.94}
\begin{table*}
         \centering
\small
\resizebox{\textwidth}{!}{
  \begin{tabular}{l @{\hskip 0.5em} c c c @{\hskip 0.5em} c c c@{\hskip 0.3em} c c c@{\hskip 0.3em} c c}
  \toprule

    & \multicolumn{2}{c}{XQuAD$_{\textsc{En}}$} & & \multicolumn{2}{c}{ASR} & & \multicolumn{2}{c}{MT}  & & \multicolumn{2}{c}{Keyboard}   \\ 
    \cmidrule{2-3} \cmidrule{5-6} \cmidrule{8-9} \cmidrule{11-12}
 Model    & EM & F1 & & EM & F1 & & EM & F1 & & EM & F1 \\
  \midrule
    \multicolumn{12}{c}{\cellcolor{light-gray} Synthetic} \\
BiDAF \cite{seo2016bidirectional} & 60.08 & 71.96 && 54.62 & 66.39 && 55.97 & 68.01 && 45.21 & 57.78 \\ 
BiDAF-ELMo \cite{peters-etal-2018-deep} & 62.61 & 75.38 && 56.81  & 70.30 &&  57.39 & 70.05  && 50.93 & 63.80 \\ 
BERT \cite{devlin2018bert}  &72.77 & 84.66 && 61.93 & 77.02 && 67.23 & 79.08 && 61.76 & 73.64\\ 
RoBERTa \cite{liu2019roberta} & 72.35 & 84.42 && 68.07 & 81.38 && 68.40 & 80.93 && 65.04 & 76.97 \\ \midrule
    \multicolumn{12}{c}{\cellcolor{light-gray} Natural} \\
BiDAF \cite{seo2016bidirectional} & 60.08 & 71.96 && 45.97 & 57.64 && 54.87 & 66.90 && 56.89 & 68.33\\
BiDAF-ELMo \cite{peters-etal-2018-deep} & 62.61 & 75.38  &&  49.16 & 62.49 && 59.24 & 71.06 && 60.76 & 73.32 \\ 
BERT \cite{devlin2018bert} & 72.77 & 84.66 && 52.94 & 67.13 && 68.82 & 79.98 && 69.16 & 81.84\\ 
RoBERTa \cite{liu2019roberta} & 72.35 & 84.42 && 60.08 & 73.61 && 70.00 & 82.13 && 70.92 & 83.37\\ 

  \bottomrule
  \end{tabular}

\caption{Performance of the QA models under the three kinds of interface noise: ASR (using Google ASR), MT (with the German--English model), and keyboard. All models score lower on noisy data, most notably on the natural ASR set. MT noise is less prominent, but we later show its impact is highly dependent on the input language.}
\label{tab:ModelPerfFirst}
}
\end{table*}

\subsection{Results and Analysis}
\label{sec:results}
\Tref{tab:cerwer} shows the character error rate (CER), word error rate (WER) and BLEU score\footnote{Uncased detokenized BLEU using SacreBLEU \cite{post-2018-call}.} %
for the generated challenge sets. Synthetic ASR and MT pipelines introduce substantially less noise than their natural counterparts, while the opposite holds for the keyboard. This is likely due to the generators not being equally controllable: while we can arbitrarily make the synthetic keyboard set noisier by increasing the corruption rate, synthetic ASR and MT pipelines include black-box components which also make the task easier for the interface by design (TTS synthesizes idealized speech, back-translation mimics MT training conditions).

In this section, we investigate how robust QA models are to these interface errors. \Tref{tab:ModelPerfFirst} reports the performance on both synthetic and natural challenge sets. For brevity, we present results using the German--English model and the Google ASR for MT and ASR respectively. 

\begin{figure}[t]
\centering
\resizebox{\columnwidth}{!}{
    \begin{tikzpicture}
      \pgfplotsset{
          scale only axis
      }
      \begin{axis}[
        xlabel=Query language,
        ylabel=F1 score,
        xtick={0, 1, 2, 3, 4, 5, 6, 7, 8, 9, 10},
        xticklabels={th, ar, zh, hi, el, tr, vi, ru, es, de, en},
        xticklabel style={text height=2ex},
        legend pos=north west,
        legend cell align={left},
        width=7cm,
        height=5cm,
      ]
        \addplot[only marks, mark=triangle*, color=blue]
        coordinates{ %
        (0, 71.03)
        (1, 75.45)
        (2, 76.39)
        (3, 76.75)
        (4, 76.96)
        (5, 76.98)
        (6, 77.14)
        (7, 78.06)
        (8, 79.86)
        (9, 79.98)
        (10, 84.66)
        }; \label{bert_langs}
        \addplot[only marks, mark=square*, color=orange]
        coordinates{
        (0, 74.68)
        (1, 76.41)
        (2, 78.88)
        (3, 79.61)
        (4, 79.6)
        (5, 79.28)
        (6, 78.77)
        (7, 79.67)
        (8, 81.65)
        (9, 82.13)
        (10, 84.42)
        }; \label{roberta_langs}
        \addlegendimage{/pgfplots/refstyle=bert_langs}
        \addlegendentry{BERT}
        \addlegendimage{/pgfplots/refstyle=roberta_langs}
        \addlegendentry{RoBERTa}
      \end{axis}
    \end{tikzpicture}
    }
    \caption{Effect of the input language on the QA system performance in an MT pipeline. Automatically translating non-English queries to English decreases performance for all source languages, and the decrease is especially noticeable for lower-resource languages.}
    \label{fig:mt-lang-variation}
\end{figure}
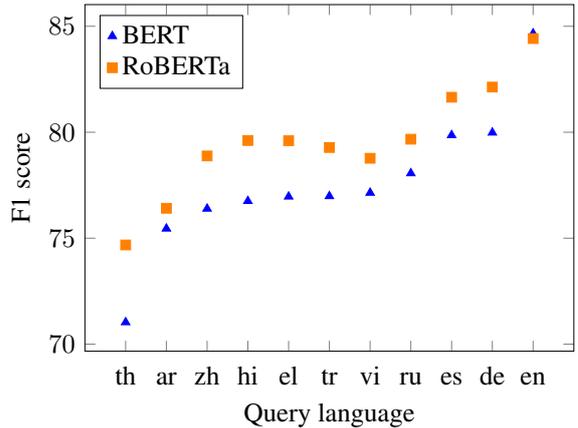

First, we observe that both synthetic and natural noise decrease accuracy for all models and interfaces, with synthetic keyboard and natural ASR errors being the most challenging. As for MT noise, \Tref{tab:ModelPerfFirst} reports results on German queries; although the systems seem robust on these, we find that MT noise can actually be quite challenging with sharp degradation of performance on Thai and Arabic (\Fref{fig:mt-lang-variation}). Further, we notice that the relative performance of models on the development set is not necessarily a sufficient proxy for the relative robustness of models to interface errors: while BERT and RoBERTa perform very similarly on XQuAD--English, RoBERTa outperforms BERT on handling all three kinds of interface errors. For practitioners, this could suggest that simply choosing the highest-accuracy QA model without separately evaluating robustness to interface noise may lead to sub-optimal performance in practice.

Below we discuss the effect of each interface in more detail.

\pgfplotstableread{
     Model   mean std
     BiDAF 53.17 12.56
     BiDAF-ELMo 52.64 13.73
     BERT 56.41 13.81
     RoBERTa 59.17 15.23
}{\mytable}

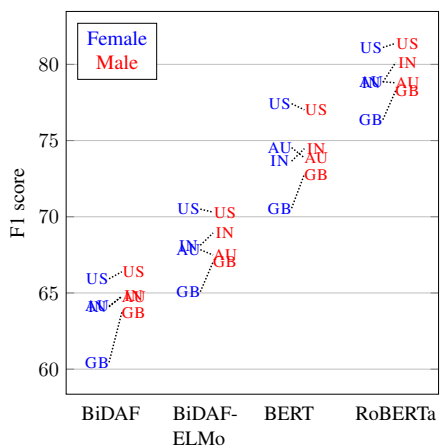
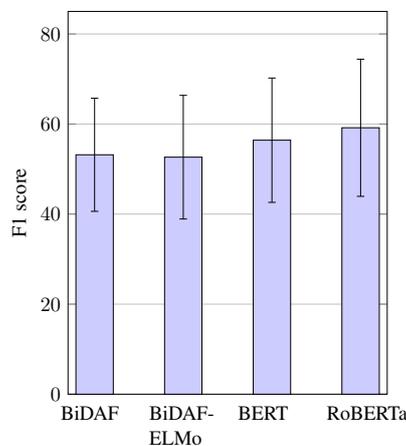
\begin{figure*}[t]
\centering
\begin{tabular}{@{}p{5.8cm}l@{}}
 \subfloat[F1 scores for TTS voices over 4 accent and 2 gender settings. Lines connect the results for different gender but same accent and model.] {
  \label{fig:synthetic_asr}
  \scalebox{.7}{
    \begin{tikzpicture}
      \pgfplotsset{
          scale only axis
      }
      \begin{axis}[
        ylabel=F1 score,
        xtick={0, 1, 2, 3},
        xticklabels={BiDAF, BiDAF-ELMo, BERT, RoBERTa},
        xticklabel style={text height=2ex, text width=0.5cm,anchor=north east},
        width=7cm,
        height=7.25cm,
        xtick style={draw=none},
        ymajorgrids=true,
        legend pos=north west,
        empty legend
      ]
        \addlegendentry[blue]{Female}
        \addlegendentry[red]{Male}
        \addplot[only marks, mark=text, text mark=\textsc{au}, color=blue]
        coordinates {
        (0-0.2, 64.14)
        (1-0.2, 67.84)
        (2-0.2, 74.54)
        (3-0.2, 78.86)
        }; \label{au_f}
        \addplot[only marks, mark=text, text mark=\textsc{gb}, color=blue]
        coordinates{
        (0-0.2, 60.45)
        (1-0.2, 65.08) 
        (2-0.2, 70.56)
        (3-0.2, 76.37)
        }; \label{gb_f}
        \addplot[only marks, mark=text, text mark=\textsc{in}, color=blue]
        coordinates{
        (0-0.2, 64.09)
        (1-0.2, 68.13) 
        (2-0.2, 73.65)
        (3-0.2, 78.83)
        }; \label{in_f}
        \addplot[only marks, mark=text, text mark=\textsc{us}, color=blue]
        coordinates{
        (0-0.2, 65.93)
        (1-0.2, 70.50)
        (2-0.2, 77.42)
        (3-0.2, 81.11)
        }; \label{us_f}
        \addplot[only marks, mark=text, text mark=\textsc{au}, color=red]
        coordinates{
        (0+0.2, 64.76)
        (1+0.2, 67.49) 
        (2+0.2, 73.87)
        (3+0.2, 78.79)
        }; \label{au_m}
        \addplot[only marks, mark=text, text mark=\textsc{gb}, color=red]
        coordinates{
        (0+0.2, 63.73)
        (1+0.2, 67.04)
        (2+0.2, 72.79)
        (3+0.2, 78.27)
        }; \label{gb_m}
        \addplot[only marks, mark=text, text mark=\textsc{in}, color=red]
        coordinates{
        (0+0.2, 64.80) 
        (1+0.2, 68.94)
        (2+0.2, 74.47)
        (3+0.2, 80.13)
        }; \label{in_m}
        \addplot[only marks, mark=text, text mark=\textsc{us}, color=red]
        coordinates{
        (0+0.2, 66.39)
        (1+0.2, 70.30)
        (2+0.2, 77.02)
        (3+0.2, 81.38)
        }; \label{us_m}
        \addplot[mark=none, color=black, thick, densely dotted]
        coordinates {(-0.07, 64.14) (0.07, 64.76)};
        \addplot[mark=none, color=black, thick, densely dotted]
        coordinates {(-0.07, 65.93) (0.07, 66.39)};
        \addplot[mark=none, color=black, thick, densely dotted]
        coordinates {(-0.07, 60.45) (0.07, 63.73)};
        \addplot[mark=none, color=black, thick, densely dotted]
        coordinates {(-0.07, 64.09) (0.07, 64.80)};
        \addplot[mark=none, color=black, thick, densely dotted]
        coordinates {(1-0.07, 67.84) (1.07, 67.49)};
        \addplot[mark=none, color=black, thick, densely dotted]
        coordinates {(1-0.07, 65.08) (1.07, 67.04)};
        \addplot[mark=none, color=black, thick, densely dotted]
        coordinates {(1-0.07, 68.13) (1.07, 68.94)};
        \addplot[mark=none, color=black, thick, densely dotted]
        coordinates {(1-0.07, 70.50) (1.07, 70.30)};
        \addplot[mark=none, color=black, thick, densely dotted]
        coordinates {(2-0.07, 74.54) (2.07, 73.87)};
        \addplot[mark=none, color=black, thick, densely dotted]
        coordinates {(2-0.07, 73.65) (2.07, 74.47)};
        \addplot[mark=none, color=black, thick, densely dotted]
        coordinates {(2-0.07, 70.56) (2.07, 72.79)};
        \addplot[mark=none, color=black, thick, densely dotted]
        coordinates {(2-0.07, 77.42) (2.07, 77.02)};
        \addplot[mark=none, color=black, thick, densely dotted]
        coordinates {(3-0.07, 78.86) (3.07, 78.79)};
        \addplot[mark=none, color=black, thick, densely dotted]
        coordinates {(3-0.07, 76.37) (3.07, 78.27)};
        \addplot[mark=none, color=black, thick, densely dotted]
        coordinates {(3-0.07, 78.83) (3.07, 80.13)};
        \addplot[mark=none, color=black, thick, densely dotted]
        coordinates {(3-0.07, 81.11) (3.07, 81.38)};
      \end{axis}
    \end{tikzpicture}
  }
  } &
  \subfloat[F1 mean and standard deviation over 4 human voices. For details and score breakdown by speaker, see Appendix~\ref{sec:appendix_asr}.] {
  \label{fig:natural_asr}
  \scalebox{.7}{
    \begin{tikzpicture}
      \pgfplotsset{
          scale only axis
      }
        \begin{axis} [
            ylabel=F1 score,
            ymin=0, ymax=85,
            symbolic x coords={BiDAF,BiDAF-ELMo,BERT,RoBERTa},
            xtick=data,
            xticklabel style={text height=2ex, text width=0.5cm, anchor=north east},
            ybar=0pt,
            bar width=20pt,
            width=6cm,
            height=7.25cm,
            xtick align=inside,
            xtick style={draw=none},
            ymajorgrids=true
        ]
        \addplot[fill=blue!20!white] 
          plot [error bars/.cd, y dir=both, y explicit, error bar style={line width=0.5pt}]
          table [y error plus=std, y error minus=std] {\mytable};
        \end{axis} 
    \end{tikzpicture}
  }
  } 
  \end{tabular}\hfill
  \begin{minipage}{0.25\textwidth}
    \caption{Effect of synthetic (a) and natural (b) voice variation on the QA performance in an ASR pipeline. Synthetic voice variation is achieved by varying accent and gender settings in the Google TTS model; US accent setting shows the highest scores while neither gender setting consistently performs best (indicated by line slopes). Natural variation is measured on a sample of 100 questions narrated by four annotators. All models exhibit considerable variation in both experiments.}
    \label{fig:tts_by_model}
    \end{minipage}
\end{figure*}

\paragraph{ASR Noise:} Speech recognizers typically omit punctuation, which could mean losing cues important for the downstream task.
To look at this factor in isolation, we remove punctuation from the original XQuAD questions. This change alone decreases BERT performance by 5.1 F1, suggesting that the absence of punctuation in part explains the degradation in the presence of ASR noise. 
When we qualitatively analyze a sample of 50 questions that BERT answered successfully in the original setting but not when passed through the speech interface, we find that 14\% of them are identical to the original modulo
punctuation. Other sources of error include the ASR producing completely meaningless questions (28\%), hallucinating (12\%) or losing named entities (10\%), and replacing words with homonyms (4\%); other difficult cases include recognizing acronyms and preserving possessives, tense, and number (2\% each). Although these problems could be diminished by designing better interfaces, we believe it is also worthwhile for practitioners to work on improving robustness of the QA system itself: many interfaces, especially commercial, only offer black-box access, and building a completely noise-free interface is not feasible.

Voice variation also plays a role: ASR error distribution differs by speaker background variables such as accent~\cite{zheng2005accent}, in turn affecting the downstream systems~\cite{accentgap, 10.1145/3308560.3317597, palanica2019you}. To emulate speaker variation in the synthetic setting, we use Google English Text-to-Speech to pronounce the XQuAD questions in eight different voices, varying the provided accent and gender settings. As \Fref{fig:synthetic_asr} shows, all models exhibit considerable variation in F1 score, consistently performing best on synthetic US accent (which our speech recognizer is optimized for) and worst on GB. Score breakdowns by setting can be found in Appendix~\ref{sec:appendix_asr}.

We 
also 
repeat the experiment with four human speakers narrating a sample of 100 XQuAD questions, to control for content. As shown in~\Fref{fig:natural_asr}, each model's performance varies substantially between voices. The 
four 
speakers differ by accent (2 Indian, 1 Russian, 1 Scottish), gender (2 male, 2 female), and level of proficiency (native and non-native); more details and individual speaker scores can be found in Appendix~\ref{sec:appendix_asr}.\footnote{Comparisons between demographics should not be drawn from per-speaker results, since we do not control for confounds like recording conditions, aiming for a realistic %
sample.} 
Although 
improving robustness to accent variation is out of the scope of our work, %
we highlight that the performance can degrade sharply depending on the user and their acoustic conditions.

We also analyze how the choice of ASR model affects the QA accuracy, focusing in particular on the decoding strategies for out-of-vocabulary words. We compare Kaldi, which outputs an UNK token for unknown words~\cite{peddinti2015jhu}, and Google's large-vocabulary ASR model. On our set of human voices, Kaldi produces at least one UNK token for $\sim$50\% of the questions, and BERT achieves an F1 score of only 43.6 on this set (54.4 F1 and 32.3 F1 separately on questions with and without UNK respectively) compared to 67.1 F1 achieved by Google ASR,
demonstrating that speech recognizer choice can greatly affect downstream QA performance.
The observed degradation due to UNK decoding~\citep[previously noted by][]{peskov2019mitigating} suggests that practitioners might find it useful to go beyond speech recognition benchmarks, and also evaluate ASR systems in the context of downstream applications.

\begin{table*}[tb]
\small
\resizebox{\textwidth}{!}{
\setlength\tabcolsep{2pt}
\begin{tabular}{p{0.05\textwidth} p{0.38\textwidth}p{0.05 \textwidth} p{0.5 \textwidth}}
\toprule
\multicolumn{1}{l}{{Language}} & \multicolumn{1}{l}{Question} &  \multicolumn{1}{l}{Language} & \multicolumn{1}{l}{Question}  \\
\toprule
\\[-1.75ex] \multicolumn{2}{l}{\textbf{What type of Lord is Doctor Who?} } & \multicolumn{2}{l}{\textbf{When would the occupation of allies leave Rhineland?} } \\ \midrule \\[-1.75ex] 
\textbf{de}: & What \tablecolor{kind} of \tablecolor{gentleman} is Doctor Who? & \textbf{de}: & When would the \tablecolor{Allied occupation} leave \tablecolor{the} Rhineland?  \\
\textbf{zh}: & What type of \tablecolor{lord} is Doctor Who? & \textbf{zh}:& When \tablecolor{was} the Allies \tablecolor{scheduled to withdraw from} Rhineland? \\
\textbf{hi}: & What \tablecolor{kind} of \tablecolor{deity} is Doctor Who? & \textbf{hi}: & When \tablecolor{will the Rhineland be removed from the occupation of the Allied countries?} \\
\textbf{ru}: & What type of \tablecolor{overlord} is Doctor Who? & \textbf{ru}: &When \tablecolor{did the Allies intend to remove the occupation of the Rhine region?} \\
\hline \\[-1.75ex] \multicolumn{2}{l}{\textbf{Who is the chair of the IPCC? }} & \multicolumn{2}{l}{\textbf{How much food does a ctenophora eat in a day? }} \\ \midrule \\[-1.75ex]
\textbf{de}: & Who is the chair of the IPCC? & \textbf{de}:& How much food does a \tablecolor{jellyfish} eat in a day??  \\
\textbf{zh}: & Who is the \tablecolor{current chairman} of the IPCC? & \textbf{zh}: &  How much food does a \tablecolor{jellyfish} eat in a day? \\
\textbf{hi}: & Who is the \tablecolor{President} of IPCC? & \textbf{hi}: & How much food does a \tablecolor{tenophora} eat in a day?  \\
\textbf{ru}: & Who is the \tablecolor{chairman} of the IPCC? & \textbf{ru}: &  How much food does a \tablecolor{ctenophore} eat per day?\\
\bottomrule
\end{tabular}
}
\caption{Examples of translation divergences for German (de), Chinese (zh), Hindi (hi), and Russian (ru). }
\label{tab:dataex-trans}
\end{table*}

\paragraph{Translation Noise:} As \Tref{tab:ModelPerfFirst} shows, German--English translation errors affect the performance of all models, although to a lesser extent than ASR noise. However, the MT quality and, in turn, the downstream performance varies greatly depending on the source language. \Fref{fig:mt-lang-variation} shows BERT and RoBERTa F1 scores on questions translated from each of the ten XQuAD languages to English (numbers reported in Appendix~\ref{sec:appendix_mt}). While German and Spanish have the highest accuracy, lower-resource and more typologically distant languages like Arabic and Thai are far behind. On translated Thai inputs, BERT achieves only 71.0 F1, which is a 16\% drop in accuracy from the original English setting compared to 6\% for German. 

\Tref{tab:dataex-trans} shows example translations from four XQuAD languages and highlights their divergences from the original questions. Since the questions are being translated out of context, MT tends to replace important content words with ones that are semantically related but not appropriate in the given context (\emph{Lord}$\rightarrow$\emph{deity}, \emph{chair}$\rightarrow$\emph{President}, \emph{ctenophore}$\rightarrow$\emph{jellyfish}). Transliteration of technical terms and named entities is also a challenge, especially for languages written in non-Latin scripts (\emph{ctenophore}$\rightarrow$\emph{tenophora} through Hindi, \emph{Jochi}$\rightarrow$\emph{Dschötschi} through German). For further qualitative analysis, we sample 100 questions translated from Hindi which BERT fails to answer correctly despite accurately answering their English equivalent. Of these, 30\% were identified by a native speaker annotator as paraphrases of the original question that would admit the original answer. The remaining incorrect translations are due to question type shift (31\%), ungrammatical or meaningless questions (12\%), corrupted named entities (8\%) and dropped determiners (2\%; Hindi does not generally use definite articles). Some divergences also go beyond word level, e.g. 10\% of questions have semantic role inversion (\emph{What earlier market did the Grainger Market replace?}$\rightarrow$\emph{Which earlier market replaced Granger's market?}). While some word-level errors can be corrected post-hoc, repairing syntax is much more challenging, which again brings it down to the robustness of the QA engine.

\definecolor{light-gray}{gray}{1.0}
\begin{table*}
\centering
\small
\resizebox{0.85\textwidth}{!}{
  \begin{tabular}{l @{\hskip 0.5em} c c c @{\hskip 0.2em} c c c@{\hskip 0.2em} c c c@{\hskip 0.2em} c c}
  \toprule

& \multicolumn{2}{c}{XQuAD$_{\textsc{En}}$} & & \multicolumn{2}{c}{ASR} & & \multicolumn{2}{c}{MT}  & & \multicolumn{2}{c}{Keyboard}   \\ 
    \cmidrule{2-3} \cmidrule{5-6} \cmidrule{8-9} \cmidrule{11-12}
BERT Model    & EM & F1 & & EM & F1 & & EM & F1 & & EM & F1 \\
  \midrule

BERT & 72.77 & 84.66 && 52.94 & 67.13 && 68.82 & 79.98 && 69.16 & 81.84\\ 
+ Named entity repair & \textbf{72.94} & \textbf{84.78}  && 53.03 & 67.34 && 68.82 & 80.05 && 69.58 & 82.22 \\
+ Content word repair & \textbf{72.94} & 84.77 && 52.61 & 67.01 && 68.32 & 79.76 && 70.25 & \textbf{82.60} \\
+ Augmentation & 72.35 & 83.89 && \textbf{64.37} & \textbf{75.89} && \textbf{68.90} & \textbf{80.83} && \textbf{70.76} & 82.43  \\
\bottomrule
\end{tabular}
}
\caption{Effect of question repair and data augmentation on BERT performance on three types of natural noise. %
Results on synthetic noise and data augmentation score breakdown by interface can be found in Appendix~\ref{sec:appendix_mitigation}.}
\label{tab:ModelPerfRobust}
\end{table*}

\paragraph{Keyboard Noise:} Synthetic keyboard noise produced by our key-swap typo generator has a much stronger effect on the QA performance than natural noise (11.1 F1 and 2.4 F1 drop respectively). We attribute this to differences in the perturbation intensity: $\sim$25\% of question words are corrupted in the synthetic setting, but only $\sim$9\% of words are corrupted under natural conditions.\footnote{Synthetic data corruption rate is a design decision and can be made to simulate the expected natural noise or be more challenging as a stress test, depending on practitioner's goals.} Interestingly, BiDAF- and BERT-based models consistently show comparable decreases in F1 score, suggesting that character-level tokenization of the former does not on its own guarantee robustness to typos.

Another factor that could affect downstream performance is error placement.
We evaluate BERT on three additional synthetic sets, introducing noise to only function words (conjunctions, pronouns, articles), only content words (which we limit to nouns and adjectives), or only commonly misspelled words (using the Wikipedia misspellings list as described in \Sref{sec:keyboard}). Synthetically perturbing all function words and all content words decreases F1 score by 6.7 and 11.7 respectively, confirming that not all words are equally important for the model finding the correct answer. Injecting the interface errors from Wikipedia into the 2,716 questions containing at least one commonly misspelled word yields F1 score of 78.6 (6.1 F1 drop), showcasing the decreased performance we would likely see in real-life user interactions.

\subsection{Mitigation Strategies}
\label{sec:mitigation}

We experiment with two strategies for improving the QA system robustness: repairing the question errors using the provided context and retraining QA models on the data augmented with synthetic noise. 
Question repair assumes availability of context,
making it unsuitable for open-domain QA, but reasonable for use cases like QA over manuals or policies \cite{feng2015applying, harkous2018polisis, ravichander-etal-2019-question}. This approach treats words that occur in the question but not the context as potential noise, attempting to replace them with the closest candidate from the context paragraph. We use character error rate as the distance metric, empirically setting the threshold to 0.5 using the synthetic set. We perform two experiments, applying the repair either only to content words (here, nouns and adjectives) or only to named entities in both the context and the question. \Tref{tab:ModelPerfRobust} shows how these repairs affect BERT performance on three types of natural noise. Named entity repair yields marginal improvements across the board, while content word repair has a stronger effect but only for keyboard errors. The proposed strategy could also be combined with other deterministic or off-the-shelf repair methods, such as adding final question marks for ASR (+6.52 F1) or using a spellchecker for keyboard (+1.41 F1).

For data augmentation, we use our synthetic noise generators to inject noise into $\sim$90K SQuAD training questions and retrain BERT on the combined clean and noisy data. As \Tref{tab:ModelPerfRobust} shows, augmentation yields improvements on all three types of natural noise over BERT trained on clean data only, but the performance of the augmented models drops slightly on the clean data. Best results on natural ASR and MT noise are obtained when the data is augmented with the same type of synthetic noise; interestingly, this is not true for keyboard noise, where ASR augmentation also works best. Although our results are preliminary, they suggest that augmentation could prove useful in enabling effective question answering in the real world.

To better understand where ASR and MT augmentation helps, we compare the performance of augmented and baseline BERT on additional challenge sets, synthesizing some common noise artifacts in isolation. We find that ASR noise augmentation improves robustness to omission of punctuation: ASR-augmented model achieves 82.7 F1 on questions with no punctuation and 82.9 F1 on questions without the final question mark (compared to 79.2 and 79.6 F1 for the baseline). Following the definitions in \Sref{sec:results}, we also experiment with removal of function and content words: both augmented models outperform baseline when all function words are dropped (76.1 F1 for ASR, and 70.2 F1 for MT, and 67.8 F1 for baseline), and ASR augmentation helps when all content words are dropped (68.6 F1 vs.\ 66.0 F1 for baseline).  Finally, we replace one randomly sampled named entity (of type LOC, ORG, or PER) per question with a placeholder, and the performance of ASR-augmented BERT drops less than that of the baseline BERT (by 2.3\% and 3.2\% respectively). This analysis suggests that ASR augmentation can make models more robust to errors in punctuation, named entities, and content words, and both ASR and MT could help with function word errors.

\paragraph{On the utility of synthetic challenge sets:}
We advocate that dataset designers always obtain natural data (with natural noise) when possible. However, in circumstances where collecting natural data is difficult, synthetic data can be useful when reasonably constructed. While the distribution of errors in our synthetically generated challenge sets differs from that in the natural ones (\Tref{tab:cerwer}), we find that the model performance ranking is consistent across all types of noise (\Tref{tab:ModelPerfFirst}), showing that synthetic noise sets could act as a proxy for model selection. Moreover, augmenting training data with synthetic noise improves model robustness to natural noise for all noise types in this study (\Tref{tab:ModelPerfRobust}), 
suggesting that synthetic noise generators may be capturing some aspects of natural noise. Our proposed generators could serve as templates for synthesizing interface noise when collecting natural data is infeasible, but individual practitioners should carefully identify and simulate the likely sources of error appropriate for their applications.

\section{Related Work}

\paragraph{Question Answering}
QA systems have a rich history in NLP, with early successes in domain-specific applications \cite{green1961baseball, Woods1977LunarRI, wilensky-etal-1988-berkeley, hirschman2001natural}. Considerable research effort has been devoted to collecting datasets to support a wider variety of applications \cite{quaresma2005question, monroy2009nlp, feng2015applying, liu2015predicting, nguyen-2019-question, jin-etal-2019-pubmedqa} and improving model performance on them 
\cite{lally2017watsonpaths, wang2018r, yu2018qanet, yang-etal-2019-end-end-open}. We too focus on QA systems but center the utility to users rather than new applications or techniques.

There has also been interest in studying the interaction between speech and QA systems. \citet{Lee_ODSQA} examine transcription errors for Chinese QA, and \citet{Lee_2018} propose Spoken SQuAD, with spoken contexts and text-based questions, but they address a fundamentally different use case of searching through speech. Closest to our work is that of \citet{peskov2019mitigating}, which studies mitigating ASR errors in QA, assuming white-box access to the ASR systems. Most such work automatically generates and transcribes speech using TTS--ASR pipelines, similar to how our synthetic set is constructed. However, our results show that TTS does not realistically replicate human voice variation. Besides, stakeholders relying on commercial transcription services will not have white-box access to ASR; our post-hoc mitigation strategies would be better suited for such cases.

\mypar{Challenge sets}
Model robustness evaluation with adversarial schemes is common in 
NLP tasks \cite{smith2012adversarial}, including dependency parsing \cite{rimell-etal-2009-unbounded}, information extraction \cite{schneider-etal-2017-analysing}, natural language inference \cite{marelli-etal-2014-sick, naik-etal-2018-stress, glockner-etal-2018-breaking}, machine translation \cite{isabelle-etal-2017-challenge, belinkov2017synthetic, bawden-etal-2018-evaluating, burlot-yvon-2017-evaluating} and QA \cite{jia-liang-2017-adversarial, aspillaga-etal-2020-stress}. Unlike most prior work, we do not create our challenge sets to break QA systems,
but rather for a more realistic evaluation of the systems' real-world utility.

\section{Conclusion}

In this work, we advocate for QA evaluations that reflect challenges associated with real-world use. %
In particular, we focus on questions that are written in another language, spoken, or typed, and the noise introduced into them by the corresponding interface (machine translation, speech recognition, or keyboard).
We analyze the effect of synthetic and natural noise in each interface and find that 
these errors can be diverse, nuanced, and challenging for traditional QA systems. Although we present an initial exploration of mitigation strategies, our primary contribution lies not in the specific challenge sets we construct or in developing new algorithms, but rather in identifying and describing 
one class of problems that practical QA systems must consider and providing a framework to measure them. We hope insights derived from our study stimulate research in making QA systems ready to face real-world users. 
We 
emphasize three considerations:

\paragraph{Sources of error:} This work studies errors introduced at the interface stage of QA pipelines. These errors are nearly ubiquitous, as users always interact with QA systems through some kind of interface. Thus, it is important for QA system designers to be mindful of distortions those might introduce. Our analysis can be extended to study the impact of interface-specific factors:  for example, how errors vary by keyboard layout (e.g.\ QWERTY vs. Dvorak or language-specific layouts like AZERTY) or preferred way of typing (e.g.\ using physical keyboards vs.\ swipe typing). Another fruitful area of study could lie in examining the accumulated impact of errors resulting from interface combinations (e.g.\ machine translation of ASR-transcribed queries) and the effects of such interface noise in languages other than English. However, interface distortion represents only one source of error that occurs in practical deployment, and future research would study further sources of variation such as how users may adapt their questions according to the interface used.

\paragraph{Context-driven evaluation:} This work focuses on practical evaluation of QA systems that takes into account the challenges associated with their real-world deployment. We hope to encourage the development of future user-centered or participatory design approaches to building QA datasets and evaluations, where practitioners work with potential users to understand user requirements and the contexts in which systems are used in practice.

\paragraph{Community priorities for QA systems:} 
While leaderboards on established benchmarks have facilitated rapid progress \cite{rajpurkar-etal-2016-squad, rajpurkar-etal-2018-know} and bolstered development of a variety of semantic models \cite{xiong2018dcn,liu2018stochastic,huang2018fusionnet, devlin2018bert}, 
we call for practitioners to consider the orthogonal direction of \emph{system utility} in their model design.  We believe these subareas to be complementary, and community attention towards both will help produce NLP systems that are both accurate and \emph{usable}.

\section*{Acknowledgments}
We thank Aakanksha Naik, Sujeath Pareddy, Taylor Berg-Kirkpatrick, and Matthew Gormley for helpful discussion and the anonymous reviewers for their valuable feedback. This work used the Bridges system, which is supported by NSF award number ACI-1445606, at the Pittsburgh Supercomputing Center (PSC). This research was supported in part by grants from the National Science Foundation Secure and Trustworthy Computing program (CNS-1330596, CNS-15-13957, CNS-1801316, CNS-1914486) and the DARPA KAIROS program from the Air Force Research Laboratory under agreement number FA8750-19-2-0200. The U.S. Government is authorized to reproduce and distribute reprints for Governmental purposes not withstanding any copyright notation there on. The views and conclusions contained herein are those of the authors and should not be interpreted as necessarily representing the official policies or endorsements, either expressed or implied, of the Air Force Research Laboratory or the U.S. Government.

\bibliographystyle{acl_natbib}
\bibliography{anthology,emnlp2020}

\clearpage
\appendix
\input{appendix}

\end{document}

%% file: appendix.tex
\section{Reproducibility details of models}
\label{sec:appendix_hyperpar}
We use the pre-trained AllenNLP implementations of BiDAF and BiDAF-ELMo\footnote{\rurl{github.com/allenai/allennlp-hub}} \cite{gardner-etal-2018-allennlp} and the HuggingFace implementation of BERT.\footnote{\rurl{github.com/huggingface/pytorch-transformers}}
We fine-tune BERT and RoBERTa on SQuAD with a learning rate of $3\mathrm{e}{-5}$ for 2 epochs, with a maximum sequence length of 384.
All models achieve good performance on the SQuAD dataset. Our trained models achieve the following F1 scores on SQuAD development set: BiDAF: 77.82, BiDAF-ELMo: 80.68, BERT: 88.75, RoBERTa: 89.93.

\section{Keyboard noise in the wild}
Common examples of keyboard typos include replacing a character with the one corresponding to an adjacent key \emph{(frame$\rightarrow$framd)}, inserting or deleting characters \emph{(between$\rightarrow$betwen, agency$\rightarrow$agenchy)}, and swapping adjacent characters within words \emph{(beroids$\rightarrow$beriods)}. Such errors exist even in textual QA datasets collected in relatively controlled settings: for example, all the error examples above actually occur in SQuAD. 
In a real-life situation of information need, where the user produces the question without being exposed to the context and the answer, these errors will likely be even more pervasive. We qualitatively analyze a sample from a dataset of questions collected from the Yahoo!\ Answers platform~\cite{miao2010automatically}, randomly selecting 50 questions from each topic (Science, Internet, and Hardware). We manually identify non-standard spellings and discard ones that are intentional, such as slang \emph{(thanks$\rightarrow$thanx)} or expression of emotion \emph{(so$\rightarrow$sooo)}. Since we are specifically interested in the errors that happen in the process of typing, we also separate out errors that could have originated in the user's mind; for example, the most frequent class of errors is omission or insertion of apostrophes in contractions, possessives and plurals, but all of them could plausibly be explained by the user's intention. Other common error types we find are incorrect whitespace placement and character substitutions (mostly plausible human errors), and character insertions, deletions or swapping adjacent characters within words (mostly interface errors); statistics and error examples can be found in \Tref{tab:yahoo}.

\begin{table}[t]
    \centering
    \resizebox{\columnwidth}{!}{
    \begin{tabular}{lll}
    \toprule
    Error type & Examples & \#Errors \\
    \midrule
    Apostrophe  & \emph{it's$\rightarrow$its, devices$\rightarrow$device's} & 55 \errorcolor{(0)} \\
    Whitespace  & \emph{anyone$\rightarrow$any one, a lot$\rightarrow$alot}& 18 \errorcolor{(4)} \\
    Deletion  & \emph{too$\rightarrow$to, \errorcolor{school$\rightarrow$schol}} & 18 \errorcolor{(10)} \\
    Substitution  & \emph{warranty$\rightarrow$warrenty, \errorcolor{will$\rightarrow$well}} & 12 \errorcolor{(2)} \\
    AdjSwap  &  \emph{\errorcolor{type$\rightarrow$tpye}, piece$\rightarrow$peice} & 11 \errorcolor{(9)}\\
    Insertion  & \emph{\errorcolor{answer$\rightarrow$asnswer}, lose$\rightarrow$loose} & \hphantom{0}9 \errorcolor{(5)} \\
    KeySwap  & \errorcolor{\emph{of$\rightarrow$if}} & \hphantom{0}1 \errorcolor{(1)} \\
    \bottomrule
    \end{tabular}
    }
    \caption{Examples of common error types observed in a manual analysis of the Yahoo!\ Answers questions. Examples identified as interface errors are highlighted in \errorcolor{blue}. \#Errors is total number of typographical errors, with \# of interface errors in parentheses.}
    \label{tab:yahoo}
\end{table}

\section{Filtering interface misspellings}
\label{sec:appendix_epitran}

Our source of human keyboard errors is the Wikipedia list of common English misspellings; some of them are likely to occur in the process of typing (e.g.\ \emph{and}$\rightarrow$\emph{adn}), while others can plausibly be explained by user misconception (e.g.\ \emph{recieve}$\rightarrow$\emph{receive}). Since our work focuses on interface errors specifically, we would like to only retain errors from the former category.

Our filtering approach is based on two assumptions: (a) interface errors must be plausible under the keyboard layout, and (b) misspellings that preserve pronunciation of the original word (e.g.\ \emph{article$\rightarrow$artical}) are more likely to be non-interface errors coming from users themselves. We use a two-step filtering heuristic: first, we retain only error categories likely to be explained by the interface noise (character deletion and insertion, adjacent character swap or adjacent key swap in QWERTY layout), and then discard spellings with similar pronunciations. Pronunciations are obtained via the Epitran G2P system~\cite{mortensen-etal-2018-epitran}, and similarity is determined by weighted edit distance.

On a sample of 100 Wikipedia misspellings manually labeled as interface or non-interface errors, the proposed heuristic shows 83\% agreement with human annotation. Applying the heuristic to the initial 4,518 word--spelling pairs, we obtain a set of 1,742 interface errors for 1,489 English words.

\section{Voice variation in ASR}
\label{sec:appendix_asr}

This section describes the details of the voice variation experiments discussed in \Sref{sec:results}. The numbers used to generate Figures \ref{fig:synthetic_asr} and \ref{fig:natural_asr} are presented in Tables \ref{tab:syntheticvarasr} and \ref{tab:naturalvarasr} respectively.

\paragraph{Synthetic variation}
We generate the synthetic voices using Google English Text-to-Speech system with four different accent settings (Australian, British, Indian, and US) and two gender settings (male and female voices). The performance of all models on these voices is presented in \Tref{tab:syntheticvarasr}. All QA models achieve highest F1 score when the questions are voiced with a US accent, which is likely explained by the ASR component being optimized for this accent specifically. Neither gender setting consistently leads to best performance across all models and accents. BiDAF and RoBERTa achieve highest scores when the US female synthetic voice is used, and BiDAF-ELMo and BERT perform best with the US male synthetic voice.

\paragraph{Natural variation}
We record the spoken versions of the 1,190 XQuAD questions voiced by three human annotators: H1 (Indian female), H2 (Russian female), and H3 (Indian male). The same three annotators and an additionally recruited annotator H4 (Scottish male) also voiced the same random sample of 100 XQuAD questions to measure the effect of voice variation in content-controlled setting.
The summary statistics (mean and standard deviation) for the sample of speakers are shown in \Fref{fig:natural_asr}, and the breakdown of each model's score by speaker is presented in \Tref{tab:naturalvarasr}.
To collect a set of recordings that is more representative of the real-life use cases, we do not control for recording conditions and other confounds, so our per-speaker results alone are not meant to be taken as evidence of the ASR or QA models being better-tuned for any of the mentioned demographics.

\begin{table*}[t]
         \centering
\resizebox{\textwidth}{!}{
  \begin{tabular}{l @{\hskip 0.5em} c c c @{\hskip 0.5em} c c c@{\hskip 0.3em} c c c@{\hskip 0.3em} c c}
  \toprule

    & \multicolumn{2}{c}{AU} & & \multicolumn{2}{c}{GB} & & \multicolumn{2}{c}{IN}  & & \multicolumn{2}{c}{US}   \\ 
    \cmidrule{2-3} \cmidrule{5-6} \cmidrule{8-9} \cmidrule{11-12}
 Model    & Female & Male & & Female & Male & & Female & Male & & Female & Male \\
  \midrule
BiDAF~\cite{seo2016bidirectional} & 64.14  & 64.76  && 60.45 & 63.73 && 64.09 & 64.80 && 65.93 & 66.39 \\ 
BiDAF-ELMo \cite{peters-etal-2018-deep} & 67.84  & 67.49 && 65.08 & 67.04 && 68.13 & 68.94 && 70.50 & 70.30 \\ 
BERT \cite{devlin2018bert} & 74.54  & 73.87 &  & 70.56  & 72.79 &  & 73.65   & 74.47 &  & 77.42   & 77.02  \\ 
RoBERTa \cite{liu2019roberta}  & 78.86 & 78.79 & & 76.37 & 78.27 & & 78.83 & 80.13 & & 81.11 & 81.38 \\
  \bottomrule
  \end{tabular}
\caption{Performance of different QA models in the TTS-ASR pipeline with different synthetic voices. We use Google Text-to-Speech with different accent and gender settings, and Google Speech-to-Text optimized for English--US as the speech recognizer.}
\label{tab:syntheticvarasr}
}
\end{table*}

\begin{table*}[t]
\resizebox{\textwidth}{!}{
\begin{tabular}{llllllllllll}  \toprule
 Model       & en & es & hi & vi & de & ar & zh & el & ru & th & tr \\ \midrule
BERT     & 84.66 & 79.86 &  76.75 &  77.14 & 79.98 & 75.45 & 76.39  & 76.96  & 78.06  & 71.03   & 76.98\\
RoBERTa & 84.42 & 81.65 & 79.61 & 78.77 & 82.13 & 76.41 & 78.88 & 79.6 & 79.67 & 74.68 & 79.28  \\ \bottomrule 
\end{tabular}
\caption{QA performance on XQuAD human translations of SQuAD questions in different source languages posed to an English QA system. Questions in each non-English language are translated to English using the Google MT system, and the performance on the original English questions is reported as a skyline.}
\label{tab:trans_alllang}
}
\end{table*}

\begin{table*}[t]
\centering
\begin{tabular}{lccccccccccc}  \toprule
 Model       & H1 & H2 & H3 & H4  \\ \midrule
BiDAF     & 58.14 & 62.86 &  31.60 &  60.07 \\
BiDAF-ELMo     & 56.15 & 62.65 &  29.30 &  62.48 \\
BERT     & 59.77 & 67.27 &  32.98 &  65.63 \\
RoBERTa & 60.74 & 74.31 & 34.14 & 67.48 & \\ \bottomrule 
\end{tabular}
\caption{Performance of the different QA models on different human annotator voices: Indian Female (H1), Russian Female (H2), Indian Male (H3), and Scottish Male (H4). We do not control for recording conditions and other confounds in this experiment, so our results are not meant to act as evidence of ASR systems being more effective for any particular demographic.}
\label{tab:naturalvarasr}
\end{table*}

\section{Input language variation in MT}
\label{sec:appendix_mt}

\Tref{tab:trans_alllang} presents the results of the query language variation experiment (\Sref{sec:results}, \Fref{fig:mt-lang-variation}). In this experiment, we use XQuAD human translation of questions into ten languages as inputs, translating them back into English through the Google Translation API. The table also reports the results on the original English SQuAD questions to serve as a skyline. As expected, lower-resource languages and languages that are more typologically divergent from English (the QA system's language) pose the biggest challenge for the MT--QA pipeline.

\section{Robustness experiments}
\label{sec:appendix_mitigation}

\Tref{tab:ModelPerfRobustSynthetic} presents the question repair and data augmentation results on both synthetic and natural noise for all interfaces. Synthetic noise sets were used for development and tuning in all experiments. \Tref{tab:ModelPerfRobustSynthetic} also breaks down data augmentation results by the specific augmentation noise source. Training on ASR noise proves helpful for natural keyboard noise as well as natural ASR noise, and robustness to natural translation noise is only improved by augmenting the data with its synthetic counterpart.

\section{ASR system benchmarking}
To benchmark both the ESPnet CommonVoice ASR system, which we use for data augmentation, and the Google ASR, which was used to create ASR challenge sets from recorded XQuAD questions, we also transcribe the natural and synthetic challenge set recordings with ESPnet ASR.
ESPnet achieves 56.8\% and 70.1\% WER for synthetic and natural voices respectively, while Google ASR gets a WER of 16.6\% and 30.7\% respectively (\Tref{tab:cerwer}).

\section{Numeral handling and ASR interfaces}

Correctly transcribing numerals is often important for producing a correct answer in an ASR--QA pipeline. Even a different representation of the same quantity in the question and in the context passage creates additional difficulties for the QA system. To additionally analyze the effect of handling numerals in ASR engines, we combine BERT with Kaldi~\cite{povey2011kaldi} or Google speech recognizers and compare their performance on the portion of XQuAD questions containing numerals (\textsc{XQuAD-numbers}) and the remaining questions (\textsc{XQuAD-nonum}). With the questions narrated by human annotators, the QA pipeline performs worse on \textsc{XQuAD-numbers} than \textsc{XQuAD-nonum} with either Kaldi (38.39 F1 and 44.30 F1 respectively) or Google ASR (64.44 F1 and 70.86 F1 respectively). In case of Kaldi, we hypothesize that the discrepancy might be partially explained by the speech recognizer outputting numbers in their spelled-out form rather than numeric form. To test this hypothesis, we convert all numerals in the original written \textsc{XQuAD-numbers} questions into their spelled-out form and observe a drop in performance from 87.10 F1 to 82.88 F1 on this subset. However, the representation mismatch is only one of many challenges: unlike Kaldi, Google ASR outputs numerals as digits, but the corresponding pipeline still shows worse performance on spoken \textsc{XQuAD-numbers}.

\definecolor{light-gray}{gray}{0.94}
\begin{table*}
         \centering
\resizebox{\textwidth}{!}{
  \begin{tabular}{l @{\hskip 0.5em} c c c @{\hskip 0.5em} c c c@{\hskip 0.3em} c c c@{\hskip 0.3em} c c}
  \toprule

& \multicolumn{2}{c}{XQuAD$_{\textsc{En}}$} & & \multicolumn{2}{c}{ASR} & & \multicolumn{2}{c}{MT}  & & \multicolumn{2}{c}{Keyboard}   \\ 
    \cmidrule{2-3} \cmidrule{5-6} \cmidrule{8-9} \cmidrule{11-12}
BERT Model    & EM & F1 & & EM & F1 & & EM & F1 & & EM & F1 \\
  \midrule
    \multicolumn{12}{c}{\cellcolor{light-gray} Synthetic} \\

BERT  &\textbf{72.77} & \textbf{84.66} && 61.93 & 77.02 && \textbf{67.23} & 79.08 && 61.68 & 74.43\\ 
+ NE Repair & 72.94 & 84.78 && 62.10 & 77.23 && 67.31 &	79.19 && 63.78 & 75.31  \\
+ Content Repair & 72.94 & 84.77 && 62.02 & 77.12 && 67.31 & 79.14 && 62.61 & 74.34  \\
+ Spelling Augmentation &72.35 & 83.89 && 56.81 & 73.68 && 65.63 & 78.09 && \textbf{67.31} & \textbf{78.83}\\ 
+ ASR Augmentation & 71.93 & 83.41 && \textbf{66.13} & 78.29 && 66.13 & 78.29 && 65.46 & 76.65 \\
+ Translation Augmentation & 70.76 & 83.17 && 61.09 & 76.42 && 66.72 & \textbf{79.70} && 59.83 & 72.29\\ 
+ Spelling+ASR+Translation Augmentation & 67.48  & 80.64  && \textbf{66.13} & \textbf{79.82} && 64.20 & 77.18 && 64.28 &  77.63 \\ \midrule
\multicolumn{12}{c}{\cellcolor{light-gray} Natural} \\ 
BERT & 72.77 & 84.66 && 52.94 & 67.13 && 68.82 & 79.98 && 69.16 & 81.84\\ 
+ NE repair & \textbf{72.94} & \textbf{84.78}  && 53.03 & 67.34 && 68.82 & 80.05 && 69.58 & 82.22 \\
+ Content repair & \textbf{72.94} & 84.77 && 52.61 & 67.01 && 68.32 & 79.76 && 70.25 & \textbf{82.60} \\
+ Spelling Augmentation & 72.35 & 83.89 && 50.84 & 66.04 && 68.49 & 80.20 && 70.25 & 82.22\\ 
+ ASR Augmentation & 71.93 & 83.41  && \textbf{64.37} & \textbf{75.89} && 68.65 & 80.32 && \textbf{70.76} & 82.43  \\
+ Translation Augmentation &70.76 & 83.17 && 53.70 & 68.11 && \textbf{68.90} & \textbf{80.83} && 68.57 & 81.05\\ 
+ Spelling+ASR+Translation Augmentation & 67.48 & 80.64 && 62.02 & 74.61 && 66.81 & 80.25 && 65.88 & 78.55 \\ 
\bottomrule
  \end{tabular}
\caption{Effect of question repair and data augmentation on BERT performance on both
synthetic and natural noise for the three interface types. Data augmentation results are presented separately for each source of training synthetic noise.
Synthetic noise sets are used for development and tuning in all experiments.}
\label{tab:ModelPerfRobustSynthetic}
}
\end{table*}